\documentclass{article}

\PassOptionsToPackage{numbers, compress}{natbib}
\usepackage[preprint]{neurips_2025}

\usepackage[utf8]{inputenc} % allow utf-8 input
\usepackage[T1]{fontenc}    % use 8-bit T1 fonts
\usepackage{url}            % simple URL typesetting
\usepackage{booktabs}       % professional-quality tables
\usepackage{amsfonts}       % blackboard math symbols
\usepackage{nicefrac}       % compact symbols for 1/2, etc.
\usepackage{microtype}      % microtypography
\usepackage{xcolor}         % colors
\usepackage{graphicx}

\usepackage{amsfonts}
\usepackage{booktabs}
\usepackage{color}
\usepackage{subcaption}
\usepackage{caption}

\usepackage{amsmath}
\usepackage{amssymb}
\usepackage{paralist}

\usepackage{multirow}
\usepackage{multicol}
\usepackage{booktabs}
\usepackage{array}
\usepackage{lipsum}
\usepackage{tabularx}
\usepackage{enumitem}
\usepackage{amsmath} 
\usepackage{enumitem}
\usepackage{pifont}
\newcommand{\cmark}{\ding{51}}%
\newcommand{\xmark}{\ding{55}}%
\definecolor{merit_red}{RGB}{255,46,99}
\definecolor{merit_dark}{RGB}{37,42,52}
\definecolor{merit_blue}{RGB}{8,217,214}
\definecolor{merit_gray}{RGB}{156,156,156}

\usepackage[pagebackref=true,breaklinks=true,letterpaper=true,colorlinks,bookmarks=false,citecolor=blue,linkcolor=blue]{hyperref}

\newcommand{\ql}[1]{\textcolor{black}{#1}}

\newcommand{\dz}[1]{\textcolor{black}{#1}}

\title{Dense360: Dense Understanding from Omnidirectional Panoramas}

\author{
      Yikang Zhou$^{1, 2}$   \quad Tao Zhang$^{1}$ \quad Dizhe Zhang$^{2}$ \quad Shunping Ji$^{1}$ \quad Xiangtai Li$^{3}$ \quad Lu Qi$^{1, 2}$ \vspace{0.3em} \\
     {\normalsize $^1$Wuhan University \quad $^2$Insta360 \quad  $^3$Peking University  } \\
    {\normalsize E-mail: zhouyik@whu.edu.cn, qqlu1992@gmail.com}
}

\begin{document}

\maketitle

\begin{abstract}
Multimodal Large Language Models (MLLMs) require comprehensive visual inputs to achieve dense understanding of the physical world. 
\dz{While existing MLLMs demonstrate impressive world understanding capabilities through limited field-of-view (FOV) visual inputs (e.g., 70°)}, we take the first step toward dense understanding from omnidirectional panoramas.
We first introduce an omnidirectional panoramas dataset featuring a comprehensive suite of reliability-scored annotations. Specifically, our dataset contains 160K panoramas with 5M dense entity-level captions, 1M unique referring expressions, and 100K entity-grounded panoramic scene descriptions.
\dz{Compared to multi-view alternatives, panoramas can provide more complete, compact, and continuous scene representations through equirectangular projections (ERP). However, the use of ERP introduces two key challenges for MLLMs: i) spatial continuity along the circle of latitude, and ii)  latitude-dependent variation in information density. We address these challenges through ERP-RoPE, a position encoding scheme specifically designed for panoramic ERP.
In addition, we introduce Dense360-Bench, the first benchmark for evaluating MLLMs on omnidirectional captioning and grounding, establishing a comprehensive framework for advancing dense visual–language understanding in panoramic settings.}

% 1. VLM对世界的Dense understanding需要关于世界的完整视觉输入.
% 2. 当powerful VLMs通过narrow-FOV (e.g., 70°)视觉输入理解世界的时候, 我们迈出了第一步in dense understanding of the world from omnidirectional panoramas. 
% 3. 我们首先Collect一个Omnidirectional Panoramas数据集, 并构建了a comprehensive suite of 带有可靠性评分的annotations. 具体的, 这个数据集包括160K全景图, 5M dense entity-level captions, 1M unique  referring expressions, 100K entity-grounded 全景场景描述. 
% 4. Omnidirectional Panoramas相比于多视图等方案具有更完整, 更紧凑, 更连续的场景信息. 但是Omnidirectional Panoramas在两方面challenge当前的VLMs: 1) 经度方向的空间循环特点在Equirectangular Projection上的体现 2) 在不同纬度信息密集不同.我们在这个工作中做出了初步尝试来解决这两个问题通过实现一种适合Omnidirectional Panoramas的位置编码, 即ERP-RoPE. 
% 5. 最后我们construct一个Dense360-Bench, which评测VLMs的全向caption和grounding能力, 来支持社区发展Omnidirectional Dense Understanding VLM.
\end{abstract}
    
\section{Introduction}
\label{sec:intro}
% will wirte introduction before sunday.

% 1. Dense Understanding for world perception是一个很重要的topic，对世界的感知有各种各样的应用
By the developments of MLLMs~\cite{liu2024llava1.5,zhu2025internvl3,bai2025qwen2_5vl,zhang2025pixel,zhang2024seeing,zhang2025enhancing,rang2025eve,han2024free,luo2024feast,wu2024controlmllm, zhou2025they}, vision-based perception tasks \ql{have evolved toward a more comprehensive and interleaved understanding}. Among these tasks, dense understanding, \ql{characterized by space-aware semantic interpretations of the physical world,} \ql{has become increasingly important in real-world applications}, including augmented and virtual reality (AR/VR)~\cite{lescop2017360}, autonomous driving~\cite{yurtsever2020survey}, and embodied artificial intelligence~\cite{chrisley2003embodied}.

% 2. 现在的方法主要focus在窄视角的，但是这种情况下很难感知全部的视角，会造成对世界的理解有偏差。比如说“"前后两个人挥手，但是我们的第一人称视角会变成对我们挥手"，阻碍了进一步的发展
Conventional methods~\cite{li2024llava,yuan2025sa2va,long-vita, ding2025pvuw, yuan20254th, zhang2024omg, zhang2023dvis, zhang2025dvis++, zhou2024improving} \ql{mainly rely on} perspective images with limited FOV, which inherently \ql{lead to} incomplete and biased understanding of \ql{densely structured environments}, \ql{especially in scenarios involving extensive spatial interactions across wide FOVs.} 
% particularly in scenarios involving extensive spatial interactions across large FOVs. 
For \ql{example}, \ql{in} a scene where two individuals \ql{are waving to} each other from opposite sides of the camera operator, a narrow perspective view \ql{may} incorrectly interpret the interaction as both individuals waving directly at the camera operator. 
\ql{Such misinterpretations significantly affect the effectiveness of dense scene understanding, thereby highlighting the need for panoramic and space-aware approaches.}
% Such issues severely restrict the effectiveness of dense understanding, motivating the need for panoramic and space-aware approaches.

% 3. intiutive解要么是video的形式以时间维度环绕一圈，要么是多视角以batch维度环绕一圈，这两种方案都有问题，总是对一个维度有所割裂，同时也会引入更大的计算负担
\ql{An intuitive solution for enhancing global scene understanding is to feed MLLMs with multi-view images, either by recording video sequences using a single moving camera or by capturing single-shot views with multiple spatially distributed cameras. However, those methods such as BEVFormer~\cite{li2024bevformer} require explicit or implicit feature fusion across these multi-view images, which introduces substantial computational overhead and limits their practicality in real-world deployments. One question raised: \textit{Can we provide MLLMs with a compact input-level representation, rather than relying on fusion at the feature level?}}
% The intuitive solution is to input the MLLMs with the several multi-view images. There are two intuitive solutions to overcome the above issues. The first approach employs video recordings to incorporate temporal information from different viewing angles. The second strategy expand the batch dimension by utilizing multiple separate cameras. However, both methods exhibit significant drawbacks by introducing incontinuity in either temporal or spatial dimension which will consequently lead to incomplete scene understanding. Moreover, these approaches will substantially increase the computational complexity.

% 4.引出全景图，更紧凑，也能同时兼顾时间和空间；但是现在尴尬的是，没有对全景图很好的探索，无论是在数据还是模型层面上，一个是少数据，一个是少模型理解，balabala
\ql{Inspired by the omnidirectional representation used in 360-degree scene understanding, equirectangular projection (ERP) images provide an ideal input format to address the aforementioned limitations. }
% Specifically, }
% In contrast, 360-degree ERP images offer an ideal representation to address the aforementioned limitations.
% \ql{Compared with heuristic mosaic strategy, .}
% As the most popular panoramic projection model, 
ERP encodes a comprehensive spatial context in a single compact format, facilitating simultaneous temporal and spatial reasoning. However, research for this area remains underexplored, with notable deficiencies in both datasets and model understanding.

\begin{figure}[t]
  \centering
  \includegraphics[width=1.0\textwidth]{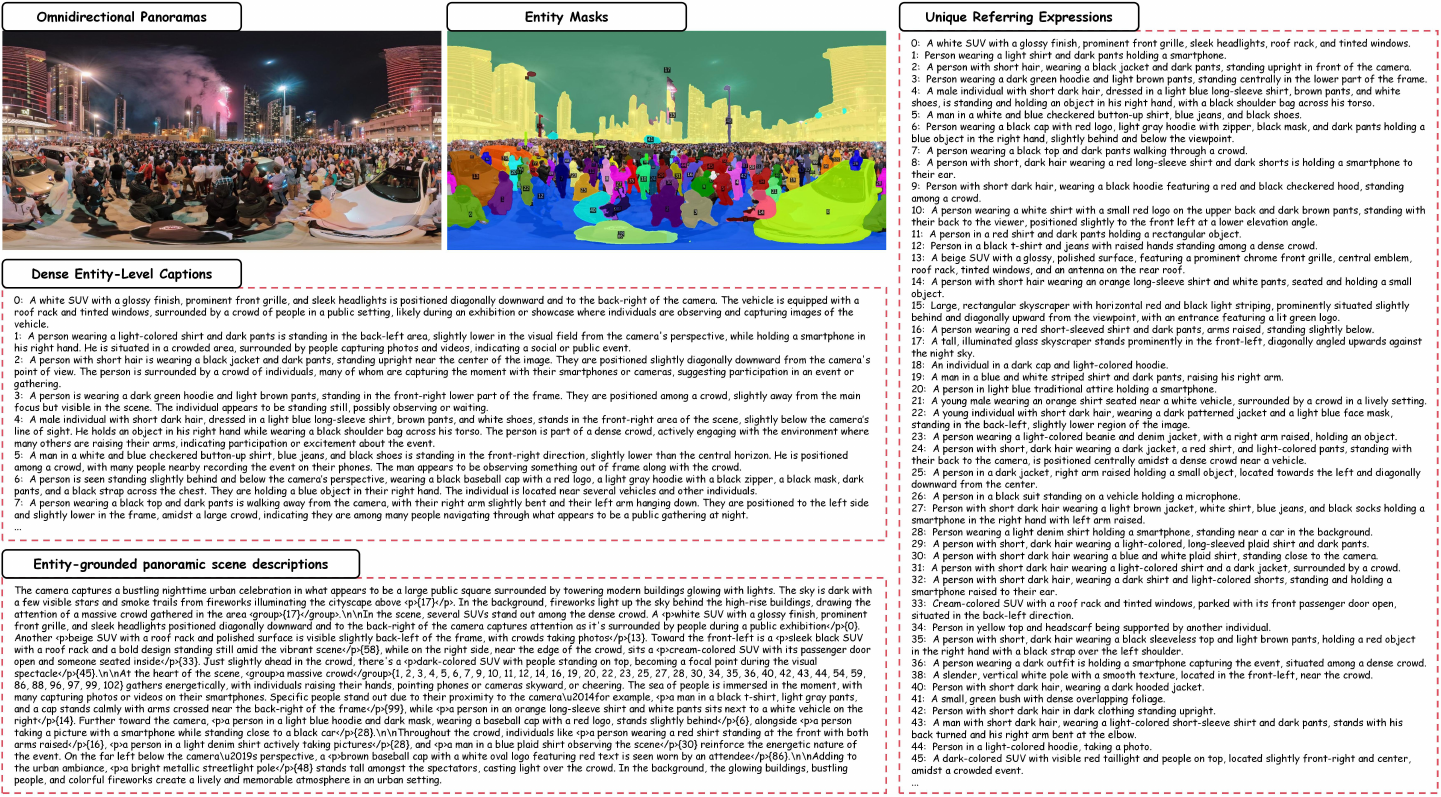}
  \vspace{-6mm}
  \caption{An Example from the Dense360 Dataset. We employ equirectangular projection (ERP) to represent omnidirectional panoramas. For each ERP image, there exists an entity-grounded panoramic scene description. The entities within ERP images are annotated with dense entity-level captions and unique referring expressions. }
  \label{fig:data_example}
\end{figure}

% 5. 为了解决这个问题，我们首先提出了一个从透视投影的模型转换为全景理解的的pipeline, 这一套pipeline能够有效的balabala，产生数据量多少balabala
\dz{To address these challenges, we propose a three-tier pipeline specific for panoramic dense understanding models. Specifically, this method leverages the visual semantics learned from the perspective domain while incorporating ERP-specific geometric priors. As a result, our pipeline enables the creation of a large-scale panoramic dataset comprising 160K panoramas with 5M dense entity-level captions, 1M unique referring expressions, and 100K entity-grounded scene descriptions (as shown in Fig.~\ref{fig:data_example}).}

% 6. 基于这批数据，我们做了一个benchmark，根据benchmark的表现，我们进一步结合全景图的特性，提出了一个更好的理解方法，balaba
\dz{Based on the proposed dataset, we further establish Dense360-Bench, a benchmark designed to evaluate MLLMs on omnidirectional captioning and grounding. Benchmarking existing MLLMs on Dense360-Bench reveals significant performance gaps compared to perspective scenarios, highlighting the limitations of conventional methods~\cite{zhu2025internvl3,bai2025qwen2_5vl} in ERP-specific dense understanding. Motivated by these challenges, we propose Dense360VLM, a vision–language framework with ERP-aware positional encoding, which achieves significant improvements for omnidirectional dense understanding.}

% 7. 我们的contribution总结如下, balabala
\dz{In summary, our contributions are threefold:
\begin{itemize}
\item \textbf{Dataset.} We introduce the largest omnidirectional dense understanding dataset to date, featuring 160K panoramas with dense, reliability-scored annotations, supporting comprehensive visual–language understanding.
\item \textbf{Benchmark.} We establish \textbf{Dense360-Bench}, the first benchmark for evaluating and advancing research on MLLMs in omnidirectional captioning and grounding tasks.
\item \textbf{Method.} We propose \textbf{ERP-RoPE}, a positional encoding scheme specifically designed for ERP representations, adapting MLLMs to the geometric characteristics of panoramic ERP.
\end{itemize}}

\dz{Together, these contributions lay a solid foundation for advancing dense visual–language understanding in omnidirectional scenarios.}

% An Example of Dense360 Dataset. 我们用equirectangular projections (ERP)来呈现omnidirectional panoramas. 对每一张ERP image有一个entity-grounded panoramic scene description。对ERP image中的entities，有dense entity-level captions和unique referring expressions。
\section{Related Work}
\label{sec:related_work}

% 1. Vision-Language Models 

% 2. Dense scene understanding

% 3. Panoramas Datasets

\noindent
\textbf{Multimodal Large Language Models. } The remarkable advancements in MLLMs~\cite{Liu2023llava,liu2024llava1.5,dai2023instructblip,chen2024internvl,wang2024qwen2,yang2023dawn,liu2024vmamba,shen2025vlm,bai2025qwen2_5vl,zhu2025internvl3, fei2025path} have demonstrated strong performance across various visual perception and understanding tasks~\cite{lin2022egocentric,huang2025egocentric,chatterjee2023opening,zhang2025mem2ego,lee2025perspective,fan2024embodied,nie2025wmnav,fu2025scene,kundu2025probres,cai2024vip,lian2025describe,lim2025ureca,lai2024lisa,liu2025seg,wang2024llm}. 
CLIP~\cite{radford2021learning}, BLIP~\cite{li2022blip,li2023blip}, ALIGN~\cite{jia2021scaling}, and other models~\cite{zeng2023clip2,zhong2022regionclip} employ contrastive learning to establish a shared embedding space between visual and textual modalities~\cite{o2015introduction,dosovitskiy2020image,brown2020language,devlin2019bert}. 
Some approaches~\cite{alayrac2022flamingo,yang2022lavt} integrate textual features into visual models or inject visual features into textual models. 
Methods~\cite{liu2023visual,dai2023instructblip,liu2024improved,li2024llava} like LLaVA~\cite{liu2023visual} utilize a projector to map visual embeddings into the feature space of large language models (LLMs). 
Building upon this paradigm, subsequent research efforts~\cite{chen2024sharegpt4v,li2024llava,liu2024improved} have focused on constructing large-scale, high-quality instruction-following datasets for model pretraining and fine-tuning. 
Recent works~\cite{zhang2025pixel,chen2024single,diao2024unveiling,diao2025evev2,luo2024mono} such as PixelSAIL~\cite{zhang2025pixel} employ a single transformer architecture as a unified vision-language model. These methodologies eliminate the dedicated vision encoder and conduct joint co-training of visual and linguistic tokens on extensive multimodal datasets.
Our work follows the Vision Encoder-Projector-LLM paradigm and introduces ERP-RoPE, a novel position encoding scheme specifically designed to handle omnidirectional panoramic inputs.

\noindent
\textbf{Panoramas Datasets.}
KITTI-360~\cite{liao2022kitti} has released a collection of urban panoramic images. 
EGOK360~\cite{bhandari2020egok360} provides an egocentric 360° kinetic human activity video dataset.
% 
% PanoVOS~\cite{yan2024panovos}  has introduced a panoramic video dataset focusing on video object segmentation tasks.
% % 
% JRDB-PanoTrack~\cite{le2024jrdb} has also presented a panoramic video dataset that contains manually-labeled panoptic segmentation and tracking annotations.
% 
PanoVOS~\cite{yan2024panovos} and JRDB-PanoTrack~\cite{le2024jrdb} introduce panoramic video datasets focusing on video object segmentation tasks.
The dataset most closely related to ours is 360+X~\cite{chen2024360+}, which is captured from multiple viewpoints with multiple data modalities. However, it primarily focuses on tasks such as scene classification and action localization. Our Dense360 dataset supports dense scene understanding through a comprehensive suite of reliability-scored annotations.

% 1. KITTI-360发布了a collection of urban panoramic images。EGOK360提供了 Egocentric 360° Kinetic human activity video dataset。PanoVOS提出了一个 panoramic video dataset focus on 视频目标分割任务。JRDB-PanoTrack也提出了一个panoramic video dataset contains manually-labeled panoptic segmentation and tracking annotations。与我们最相近的是360+X数据，which is captured from multiple viewpoints with multiple data modalities. 但是它主要关注场景分类和Action Localisation等任务。我们的Dense360数据集支持dense scene understanding by a comprehensive suite of reliability-scored annotations。

% 1. PanoVOS
% 2. 360+X
% 3. 其他全景数据集【JRDB】在纬度方向的视角范围有限，关注基础的感知任务（分割，追踪）
\section{Dens360 Dataset and Benchmark}
\label{sec:dataset_benchmark}
% \cm{[QL: It is better for us to introduce the overall pipeline and then introduce them separately.]}

\begin{figure}[t]
  \centering
  \includegraphics[width=1.0\textwidth]{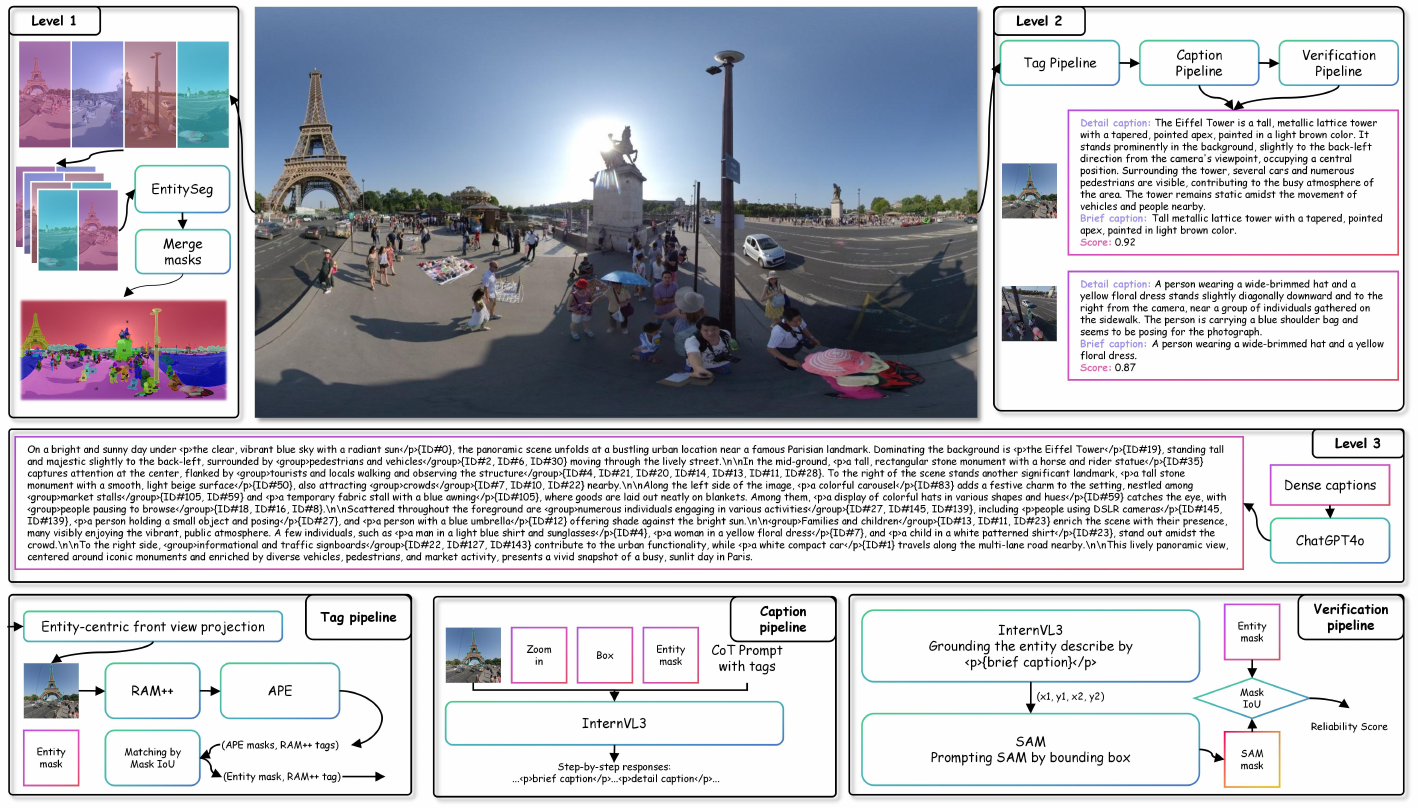}
  \vspace{-6mm}
  \caption{Dataset Generation Pipeline. The Level-1 pipeline generates entity masks. The Level-2 pipeline produces dense captions. The Level-3 pipeline creates entity-grounded panoramic scene descriptions.}
  \vspace{-5mm}
  \label{fig:data_engine}
\end{figure}

\textbf{Overall Pipeline.}
To construct an omnidirectional dense understanding dataset, we design a three-tiered pipeline, as shown in Fig.~\ref{fig:data_engine}. The Level-1 pipeline generates entity masks with granularity consistency (detailed in \S\ref{sec:level1}). The Level-2 pipeline produces dense captions for entities and assigns reliability scores to the generated captions (detailed in \S\ref{sec:level2}). The Level-3 pipeline creates entity-grounded panoramic scene descriptions for omnidirectional panoramas (detailed in \S\ref{sec:level3}).

\begin{figure}[t]
  \centering
  \includegraphics[width=1.0\textwidth]{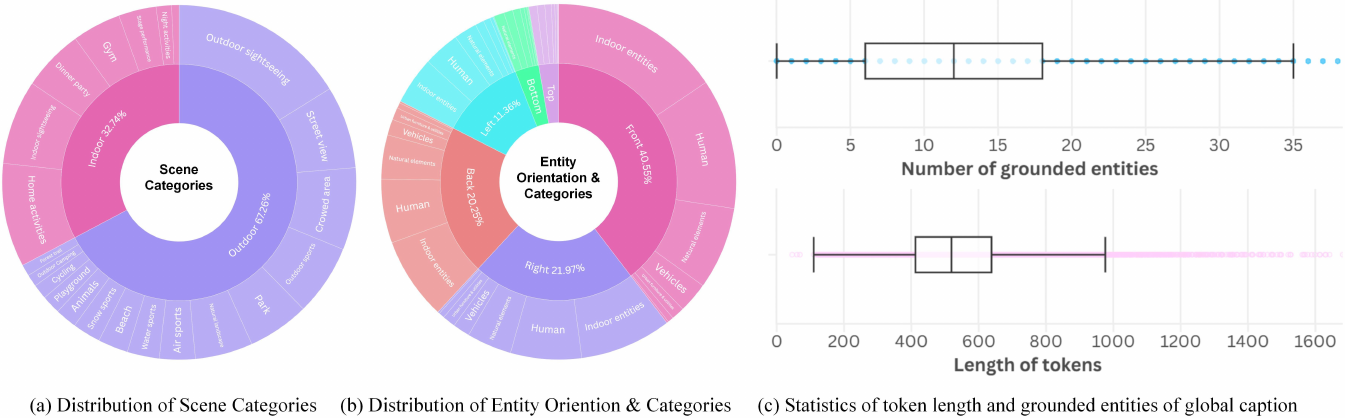}
  \vspace{-6mm}
  \caption{Dataset Statistics. We employ Qwen2.5VL-72B-Instruct~\cite{bai2025qwen2_5vl} for recognizing scene categories. We utilize Qwen2.5-72B-Instruct~\cite{yang2024qwen2} to identify entity categories from captions. We calculate length of tokens using the Qwen2.5 tokenizer~\cite{yang2024qwen2}.}
  \label{fig:data_statistics}
\end{figure}

\textbf{Dataset Statistics.}
The 160K ERP images in our Dense360 Dataset encompass diverse scene categories. 
As shown in Fig.~\ref{fig:data_statistics}(a), 32.74\% of these ERP images depict indoor scenes such as home activities, indoor sightseeing, dinner parties, and gyms, while 67.26\% represent outdoor scenes including outdoor sightseeing, street views, outdoor sports, and natural landscapes. 
We employ \ql{a} dataset generation pipeline to \ql{automatically annotate 160K ERP images}, \ql{incorporating quality control mechanisms for entity-level captions based on reliability scores.}
\ql{This process yields a large-scale corpus comprising 5 million dense, entity-level captions.}
\ql{As illustrated in Fig.~\ref{fig:data_statistics}(b), the spatial distribution of the 5 million annotated entities is uneven across the panoramic scenes: 40.55\% are located in the front quadrant, followed by 21.97\% in the right, 20.25\% in the back, and 11.36\% in the left.}
\ql{A small remaining portion is distributed across the top and bottom regions.}
\ql{Most} entities represent humans and indoor entities such as \ql{couches}, dining tables, and televisions. 
From the \ql{original set of} 5 million entities, we filter out those with low-quality masks (e.g., perforated masks, small-area masks, masks containing multiple disconnected regions) \ql{and retain a high-quality subset of} 1 million entities. 
\ql{These selected entities are then used to construct 1 million unique referring expressions based on their corresponding brief captions.}
Using the Level 3 pipeline, we generate 100K entity-grounded panoramic scene descriptions. 
As demonstrated in Fig.~\ref{fig:data_statistics}(c), within these 100K descriptions, \ql{each caption contains a median of 12 grounded entities, with a median token length of 519, indicating the fine-grained and information-dense nature of the generated annotations.}

\begin{figure}[t]
  \centering
  \includegraphics[width=1.0\textwidth]{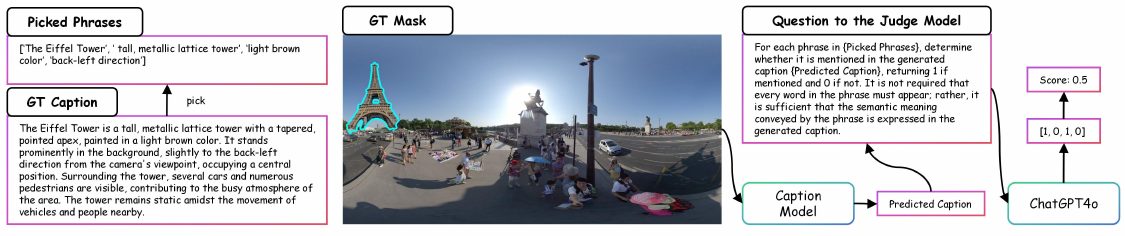}
  \vspace{-6mm}
  \caption{An example for evaluating captioning capability. The caption model refers to the MLLM under evaluation, where we employ ChatGPT-4o as the judge model.}\vspace{-5mm}
  \label{fig:eval_captioning}
\end{figure}

\textbf{Building Dense360-Bench for Grounding and Captioning.}
Grounding and captioning are two fundamental capabilities of \ql{MLLMs for} omnidirectional dense understanding. 
\ql{To evaluate these capabilities, we introduce Dense360-Bench, a benchmark designed for assessment of grounding and captioning performance in ERP images.}
\ql{From a curated set of 1,279 ERP images, we select 3,000 entities and construct 3,000 grounding questions based on their brief captions, as well as 3,000 captioning questions derived from their detailed captions.}
These 3,000 entities are evenly distributed across the front, back, left, and right \ql{panoramic directions to ensure spatial diversity.}
For grounding tasks, \ql{we follow conventional protocols by requiring MLLMs to segment the corresponding entity in the ERP image based on the brief caption.}
\ql{Thus, the performance is evaluated using the mask IoU metric.}
For captioning tasks, we design a cost-effective evaluation scheme. 
As shown in Fig.~\ref{fig:eval_captioning}, we extract key phrases \ql{from each entity’s detailed caption.}
\ql{Given} a predicted caption, we formulate \ql{a series of yes/no} questions \ql{to determine whether each key phrase is explicitly mentioned.}
We employ ChatGPT-4o as the judge model. 
\ql{Based on its responses, we construct a binary vector indicating phrase coverage and compute the recall of key phrases as the final evaluation metric.}

\subsection{Entity Masks (Level-1)}
\label{sec:level1}

At this stage, our objective is to extract granularity-consistent entities from omnidirectional ERP images. 
\ql{Compared to the traditional perspective images,} the entities in ERP images exhibit three distinctive characteristics \ql{including} high density, spatial continuity along the circle of latitude, and geometric distortion. 
% \cm{[QL: Do not understand what's the continuity along longitude.]}
%
These properties make it challenging to generate the corresponding entity masks directly from panoramic images. 
As illustrated in the Level 1 Pipeline of Fig.~\ref{fig:data_engine}, 
\ql{we begin by partitioning the ERP image into three square slice views, each with a 1:1 aspect ratio, using overlapping windows with 50\% stride to ensure spatial continuity.}
% we first partition the ERP image into three slice views with 1:1 aspect ratio using 50\% overlapping windows. 
%

\ql{To preserve spatial continuity, especially at the circular boundary where the leftmost and rightmost edges of the ERP image represent adjacent regions, we introduce a fourth slice view by stitching these opposing boundaries together.}
% To address spatial continuity where the rightmost and leftmost boundaries represent adjacent space, we additionally create a fourth-slice view by stitching these opposite boundaries. 
%
\ql{Overall, this slicing strategy has two key advantages.}
\ql{On one hand, it simplifies the densely populated panoramic scene by decomposing it into square slice views with relatively simpler content, reducing visual complexity and aligning well with the input requirements of standard segmentation models.} 
% \cm{[QL: Here, I am not convinced why square view can align with standard segmentation models. Please provide some evidence if we have time.]}
% This slicing strategy offers three advantages: First, it decomposes the densely populated scene into multiple slice views with relatively simple content.
%
% Second, the square aspect ratio aligns better with standard segmentation model inputs. 
%
% Third, the spatial continuity can be effectively restored through the high overlap ratio among the four slice views. 
\ql{On the other hand, spatial continuity across the panoramic image is effectively preserved through the high overlap ratio among the four slice views.}
%
% We also retain the original ERP image as an additional input to handle large-scale entities spanning the entire ERP image, such as floors, ceilings, and sky regions.
\ql{Additionally, we retain the original ERP image as an auxiliary  input to accommodate large-scale entities that span the entire panoramic field, such as floors, ceilings, and sky regions.}

We employ CropFormer~\cite{qi2023high} to perform entity segmentation in both the four slice views and the complete ERP image, leveraging \ql{its good consistency to image views and granularity.} 
Following entity segmentation, we project all generated masks back to the ERP view and compute a mask Intersection-over-Union (IoU) matrix across all entity instances. 
\ql{Union operations subsequently merge entity mask pairs that exhibit high IoU values (>0.7).}
This postprocessing step effectively integrates fragmented entity components into complete and granularity-consistent masks, while preserving precise boundary definitions.

\renewcommand{\arraystretch}{1.5}
\begin{table}[t]
    \centering
    \caption{Chain-of-Thought to prompt MLLMs to generate entity captions step-by-step.}
    \label{tab:cot}
    \resizebox{0.85\textwidth}{!}{
    \begin{tabular}{c c c}
    \toprule[1.5pt]
    \textbf{Reasoning step} & \textbf{Goal} & \textbf{Information source} \\
    \midrule[1pt]
    Step 1 & Assess entity mask quality & Entity mask \\
    Step 2 & Evaluate semantic clarity & Zoomed-in image \\
    Step 3 & Identify attributes & Tags \& images\\
    Step 4 & Recognize components & Zoomed-in image \\
    Step 5 & Determine Panoramic Spatial Location & Precomputed latitude and longitude \& bounding box \& contour\\
    Step 6 & Observe all entity-related events & front cube view \\
    Step 7 & Synthesize a unique caption & Previous reasoning process\\
    Step 8 & Synthesize an Entity-Centric Paragraph Caption & Previous reasoning process\\
    \bottomrule[1.5pt]
    \end{tabular}
    }
\end{table}

\subsection{Entity Caption (Level-2)}
\label{sec:level2}
At this stage, our \ql{aim} to obtain detailed captions that fully describe each entity and brief captions that capture essential information. 
\ql{Directly} using visual prompts to instruct powerful \ql{existing} MLLMs presents an off-the-shelf solution. 
However, \ql{the} MLLMs such as InternVL3, Qwen2.5VL, and ChatGPT4o struggle to accurately align with user-provided visual prompt instructions, resulting in extremely low efficiency for this approach.
\ql{Furthermore}, \ql{such a} solution lacks \ql{mechanisms for} hallucination detection \ql{in} generated captions and \ql{for verifying the alignment between captions and corresponding masks}, leading to unreliable caption outputs. 
To address these issues, we design three specialized \ql{steps}, as shown in the Level 2 Pipeline in Fig.~\ref{fig:data_engine}. 
First, a tag pipeline extracts semantic information about entities. 
Next, multiple visual prompts and a semantic-enhanced Chain-of-Thought (CoT) prompt guide MLLMs to generate both brief and detailed captions stepwise. 
Finally, a verification pipeline evaluates the reliability of the caption \ql{for proofreading}.

\textbf{Tag Pipeline.}
\noindent
% The tag pipeline integrates RAM++~\cite{huang2023open} and APE~\cite{shen2024aligning}, with the former being a recognition model and the latter a grounding model. 
\ql{The tag pipeline integrates RAM++\cite{huang2023open} as a recognition model and APE\cite{shen2024aligning} as a grounding model, enabling both category identification and spatial localization of entities.}
First, we perform a cubemap projection (CMP) on the ERP image based on the entity's location to obtain an entity-centric front view. 
% \cm{[QL: we should point out the way for this transformation such as tangent?]}
%
RAM++ takes this entity-centric front view as input and returns a set of recognized tags. 
These tags encompass diverse visual information, including entity categories, scene types, color attributes, and \ql{some} visual characteristics \ql{e.t.c}. 
\ql{Then,}
we directly prompt APE with all the tags \ql{to} generate a segmentation mask \ql{that corresponds to} each semantically meaningful tag. 
%
% At this stage, we obtain APE mask-tag pairs. 
\ql{This process yields a set of mask–tag pairs produced by APE}
Finally, we establish matching relationships by calculating the mask Intersection over Union (IoU) between the entity and APE masks. 
The tags of matched APE masks are assigned to the corresponding entity, thereby acquiring semantic prior information for the entity.

\textbf{Caption Pipeline.}
\noindent
We obtain brief captions and detailed captions by prompting InternVL3~\cite{zhu2025internvl3} with combined visual prompts and CoT prompts enhanced by prior semantic information. 
Different types of visual prompts exhibit distinct advantages and \ql{face specific} limitations. 
\ql{For example,} zoomed-in images of entities \ql{can capture} fine-grained visual details \ql{but sacrifice global contextual information.} 
Bounding boxes \ql{can provide salient} visual cues yet may introduce ambiguity when \ql{multiple entities are enclosed.}
\ql{Contour-based prompts help reduce such ambiguity but are less effective for non-compact or irregularly shaped entities.}
%
% masks enable precise entity specification but disrupt visual information. 
\ql{Mask-based prompts allow for precise entity specification, yet they may confuse surrounding visual information and disrupt perceptual coherence.}
We combine these four \ql{types of} visual prompts in a multi-image format as input to MLLM, instructing the MLLM to progressively leverage these visual prompts alongside prior semantic information through step-by-step reasoning. 
\ql{CoT prompting process is designed to progressively guide the generation of entity-centric descriptions through a structured sequence of reasoning. 
% \cm{[QL: Having a table to illustrate the steps is better.]}
%
\ql{As shown in Tab.~\ref{tab:cot}, we have eight steps in such as process.}
It begins by assessing the quality of the entity mask to ensure accurate segmentation, followed by evaluating the semantic clarity of the entity's category and identity. 
Then it identifies key visual and contextual attributes, leveraging prior tag information when available. 
Next, it recognizes relevant components associated with the entity, such as subparts, attached objects, and functional features. 
The entity's spatial location is then determined within the panoramic context, aided by previously known location data. 
To capture dynamic context, the process observes all events and interactions that involve the entity across the temporal sequence. 
Based on the accumulated information, a unique and concise caption is synthesized. 
Finally, our caption process generates a detailed, entity-centric paragraph that integrates spatial, temporal, and semantic cues into a coherent narrative.
}
%
% Our CoT prompt framework comprises eight steps:
% Step 1. Assess Entity Mask Quality;
% Step 2. Evaluate Semantic Clarity;
% Step 3. Identify Attributes (augmented by prior tag information);
% Step 4. Recognize Components including sub-parts, attached objects, and functional features;
% Step 5. Determine Panoramic Spatial Location (enhanced by prior entity location data);
% Step 6. Observe All Entity-Related Events;
% Step 7. Synthesize a Unique Caption;
% Step 8. Generate an Entity-Centric Paragraph Caption.
%
\ql{Through those steps,} we extract two distinct outputs: detailed captions that comprehensively describe each entity's characteristics and brief captions that capture essential entity information. 

\textbf{Verification Pipeline.}
\noindent
We leverage the powerful grounding capabilities \ql{of} InternVL3~\cite{zhu2025internvl3} \ql{by} instructing it \ql{to localize target entities} based on brief captions. 
\ql{The resulting grounded bounding boxes are then used to prompt SAM~\cite{ravi2024sam}, which generates segmentation masks within the specified regions.}
% We further prompt SAM~\cite{ravi2024sam} with the grounded bounding boxes to obtain masks within those regions. 
%
\ql{We note that the grounded SAM mask should be inherently aligned with the input caption, regardless of whether the caption accurately describes the intended entity.}
%
% Regardless of whether the brief caption accurately describes the wanted entity, the grounded SAM mask and the brief caption remain strictly aligned. 
%
\ql{This alignment allows us to assess the consistency between the brief caption and the target entity by comparing the SAM-generated mask with the ground-truth entity mask.}
% Therefore, we can determine the alignment between the brief caption and the entity mask by verifying whether the SAM mask aligns with the entity mask. 
%
% Naturally, the Mask IoU between the SAM mask and the entity mask can serve as a reliability score for the caption.
\ql{Therefore, the IoU between the SAM mask and the entity mask serves as a quantitative reliability score for evaluating the caption’s accuracy in grounding the correct visual region.}

\subsection{Entity-Grounded Panoramic Scene Description (Level-3)}
\label{sec:level3}
\ql{Leveraging the generated dense entity captions, we directly prompt GPT-4o to produce entity-grounded panoramic scene descriptions.}
%
% With the generated dense entity captions, we directly prompt ChatGPT4o to generate entity-grounded panoramic scene descriptions. 
%
As shown in the Level 3 Pipeline of Fig.~\ref{fig:data_engine}, \ql{these descriptions enable a comprehensive, fine-grained, and densely grounded understanding of ERP images, effectively capturing both semantic detail and spatial context.}

\section{Dense360 VLM}
\label{sec:model}

\begin{figure}[t]
  \centering
  \includegraphics[width=1.0\textwidth]{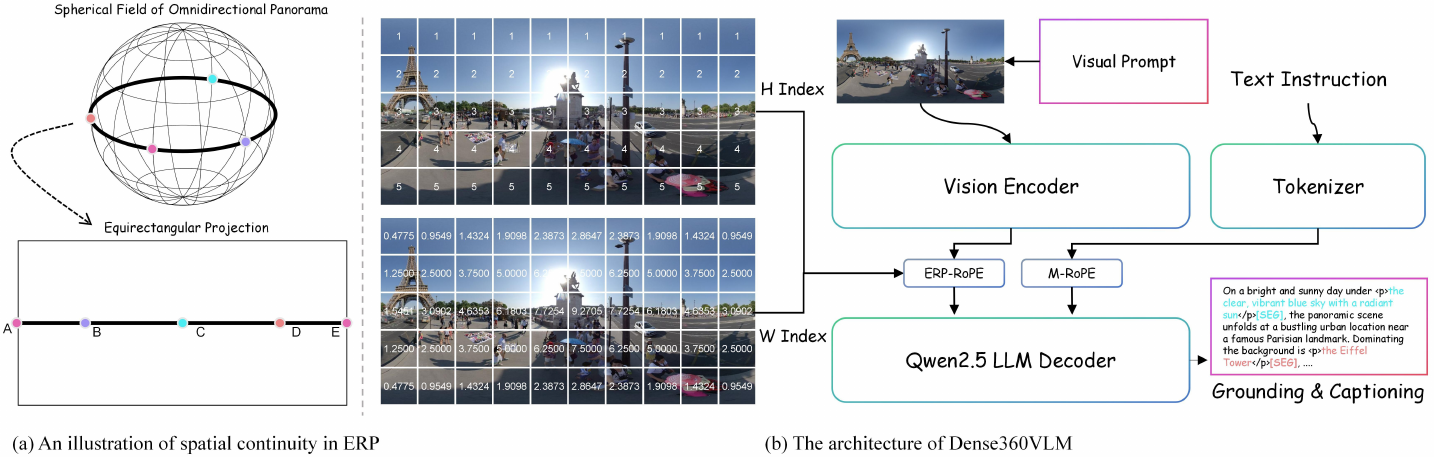}
  \vspace{-6mm}
  \caption{The architecture of Dense360VLM. The left side illustrates the relationship between pixel positions in the ERP image and their corresponding physical spatial locations. The right side demonstrates a position encoding derived using ERP-RoPE, along with the architectural framework of our Dense360VLM.}
  \label{fig:erp_rope}
\end{figure}

To enable dense visual–language understanding from panoramic ERP, we introduce \textbf{Dense360VLM}, a vision–language model tailored for ERP inputs. This section presents its two core components: the ERP-specific positional encoding \textbf{ERP-RoPE} (\S\ref{sec:erp_rope}) and its integration into the vision–language framework (\S\ref{sec:vlm}).
\subsection{ERP--RoPE}
\label{sec:erp_rope}
Unlike perspective images, Equirectangular Projection (ERP) images present two unique geometric characteristics:

\noindent\textbf{Horizontal continuity.} Since the horizontal axis of ERP corresponds to the unfolding of latitude circles on the sphere, the spatial representation in ERP images does not maintain a direct correspondence with the physical 3D environment. 
As shown in Fig.~\ref{fig:erp_rope}(a), points A and E appear farthest apart in the ERP image, yet their actual physical positions should be adjacent. The distance from point A to D in the ERP image is greater than that from A to B, although their real-world physical distances should be equal. In actual physical space, the point farthest from A should be C rather than E.

\noindent\textbf{Latitude-dependent distortion.} ERP images also introduce latitude-dependent distortion, where horizontal lines at higher latitudes are physically shorter than those at lower latitudes. Consequently, this distortion causes the pixel-based information density to decrease progressively from the equator to the poles.

\noindent Existing positional encodings---e.g.\ the rotary position embedding (RoPE) family---ignore both effects and yield sub‐optimal performance on panoramic tasks. We therefore extend multimodal RoPE (mRoPE) and propose \emph{ERP--RoPE}, a task‐agnostic positional encoding tailored for ERP inputs.

To encode positional information for ERP images, we adopt a conventional two-dimensional coordinate system.
Given a pixel located at \((h, w)\),  where \(w \in [1, W]\) and \(h \in [1, H]\), its positional encoding is formulated as \((g(h), f(w))\). 
Since the vertical coordinate \(h\) does not exhibit specific geometric distortions, we directly define \(g(h) = h\). In contrast, the horizontal coordinate \(w\), which corresponds to the latitude circles of the sphere, requires special consideration due to the inherent \textbf{horizontal continuity}.
We first summarize the mathematical properties that horizontal positional encoding in ERP images should satisfy in Tab.~\ref{tab:erp_properties}.

\renewcommand{\arraystretch}{1.5}
\begin{table}[t]
    \centering
    \caption{Mathematical properties required for ERP horizontal positional encoding.}
    \label{tab:erp_properties}
    \resizebox{0.85\textwidth}{!}{
    \begin{tabular}{c c c}
    \toprule[1.5pt]
    \textbf{Property} & \textbf{Mathematical Expression} & \textbf{Explanation} \\
    \midrule[1pt]
    Periodicity & \(f(w) = f(w + kW)\) & Ensures periodicity with a period of \(W\). \\
    Boundary consistency & \(f(w_A+1) = f(w_E)\) & Guarantees seamless continuity between the left and right borders. \\
    Symmetry & \(f(w_B) = f(w_D)\) & Maintains symmetry around the center. \\
    Maximum at ERP center & \(f(w_C) = \max(f(w)) = \frac{W}{2}\) & Ensures the center position reaches the maximum encoding value. \\
    \bottomrule[1.5pt]
    \end{tabular}
    }
\end{table}

Considering the latitude-dependent distortion in ERP images, the actual length of each latitude circle is $\cos\theta$ times the equatorial length, where $\theta \in [-90^\circ, 90^\circ]$ represents the latitude value. We introduce a scaling factor $\gamma$ to control the spacing of positional encoding for pixels, ensuring that the total length of all latitude circles after scaling equals $H \times W$:
\begin{equation}
\sum_{\theta}\cos\theta\;W\,\gamma=H\times W
\label{eq:gamma2_constraint}
\end{equation}
From this constraint, we derive the value of $\gamma$:
\begin{equation}
\gamma = \frac{H}{\sum_{\theta}\cos\theta}
\label{eq:gamma2}
\end{equation}
Finally, we reparameterize the pixel position $(h, w)$ in the ERP image as $(h, \gamma·f(w))$. For the implementation of $f(w)$, we adopt a naive solution in our code. When $W$ is an odd number:
\begin{equation}
f(w) := [1,2,3,...,(W+1)/2,(W+1)/2,(W+1)/2-1,...,3,2].index\_select(w)
\label{eq:func_w_1}
\end{equation}
When $W$ is an even number:
\begin{equation}
f(w) := [1,2,3,...,W/2+1,W/2,...,3,2].index\_select(w)
\label{eq:func_w_2}
\end{equation}
In Fig.~\ref{fig:erp_rope}(b), we visualize the computed $f(w)$ (W Index) for each ERP visual token using the above method.

\subsection{Integrate ERP-RoPE into the MLLM}
\label{sec:vlm}
We follow the Vision Encoder-Projector-LLM paradigm, using Qwen2.5VL~\cite{bai2025qwen2_5vl} as the baseline and implementing minor modifications based on it. We incorporate ERP-RoPE to provide customized positional encoding for ERP visual tokens. For other inputs such as perspective images and texts, we employ mRoPE~\cite{bai2025qwen2_5vl} to conduct positional encoding.

Although Qwen2.5VL inherently possesses grounding capability with textual bounding boxes, these bounding boxes are inadequate for addressing distortion issues and horizontal continuity challenges in ERP images. For instance, an entity located at the back of a panoramic scene would be split into two parts in an ERP image, appearing at the leftmost and rightmost ends, respectively. A viable solution~\cite{yuan2025sa2va,lai2024lisa,zhang2025pixel} involves injecting a special token ``[SEG]'' into Qwen2.5VL's vocabulary to represent entities in textual responses, which can then be decoded into corresponding masks using SAM~\cite{ravi2024sam}. Adopting this streamlined approach - integrating ERP-RoPE into Qwen2.5VL and augmenting the vocabulary with the ``[SEG]'' token - we have developed Dense360VLM to enable dense understanding of ERP image.

\section{Experiments}
\label{sec:exp}
% main exp:
% 1. caption, 这个可以测比较多的Models, 其他的Models都直接在原图上画contours.
%    可以测多个方面, 比如方位, category, 颜色, 参与的活动, 其他属性...
% 2. refseg, 这个可能不好比较, 一种方案是直接上InternVL3+SAM, Qwen2.5VL+SAM的方法
% 3. GCG, 这个更不好比较了
% (只测 caption和refseg, 这个作为main experiments. gcg 定性展示)

% Benchmark results. 我们报到了MLLMs在omnidirection和前后左右四个方位上的captioning和grounding得分。MLLM\dag表示用我们的Dense360 dataset进行了第二次的后训练。

% \begin{figure}[t]
%   \centering
%   \includegraphics[width=1.0\textwidth]{figs/inference_gcg.pdf}
%   \vspace{-6mm}
%   \caption{An entity-grounded scene description generated by Dense360VLM.}
%   \label{fig:infer_gcg}
% \end{figure}

\textbf{Evaluation Benchmark.}
\noindent
Since there is currently no benchmark to evaluate the dense understanding capabilities of MLLMs in omnidirectional panoramas, we constructed our own Dense360-Bench. We assess MLLMs' understanding of omnidirectional panoramas through entity captioning and visual grounding. For grounding, we use mask IoU as the evaluation metric; for captioning, we employ recall rate as the metric.

\textbf{Baseline Model and Training Datasets.}
\noindent
We adopt Qwen2.5VL~\cite{bai2025qwen2_5vl} as the baseline model to validate the effectiveness of our data and ERP-RoPE. The model is trained using 30\% of the LLaVA SFT data~\cite{liu2024llava1.5} and the full Dense360 dataset.

\textbf{Implementation Details.}
\noindent
We initialize Dense360VLM with the weights of Qwen2.5VL-3B-Instruct, implemented via the Xtuner codebase~\cite{2023xtuner}. The visual encoder is frozen, while the Qwen2.5 LLM decoder is fine-tuned using LoRA~\cite{hu2021lora}. We train Dense360VLM using 8 H20 GPUs.

\subsection{Main Results}
\label{sec:main_exp}

As shown in Tab.~\ref{tab:main_exp}, we report the scores of the latest ChatGPT4o, two foundational MLLMs~\cite{zhu2025internvl3,bai2025qwen2_5vl}, and two specialized small-scale expert MLLMs developed from them on Dense360-Bench. Among them, SA2VA-4B~\cite{yuan2025sa2va} is developed based on InternVL2.5~\cite{chen2024expanding}, while Dense360VLM-3B is built upon Qwen2.5VL~\cite{bai2025qwen2_5vl}. Here, we conduct a second post-training of SA2VA-4B using the Dense360 dataset to enable its comprehension of ERP image inputs. 
To test grounding capability, we first prompt Qwen2.5VL and InternVL3 to output textual grounding bounding boxes for given expressions, then use these bounding boxes to prompt SAM~\cite{ravi2024sam} for generating corresponding masks. 
This process allows calculation of the mask IoU metric, which aligns with our verification pipeline (details in \S\ref{sec:level2}). Notably, SA2VA and Dense360VLM can directly output grounding masks.

We report not only the overall scores of MLLMs on entities across all directions (omnidirection), but also their performance specifically on front, back, left, and right directions - with particular emphasis on the back direction. In ERP images, entities in the back direction are split between the extreme left and right edges. Accurate comprehension of back-direction entities most directly reflects a MLLM's spatial understanding of panoramic scenes.

The two open-source MLLMs~\cite{zhu2025internvl3,bai2025qwen2_5vl} achieve relatively low scores in both captioning and grounding, which may be attributed to their lack of training data on ERP images. The latest ChatGPT4o demonstrates exceptional panoramic scene understanding capabilities, attaining a zero-shot captioning score of 46.42 that approaches the performance of in-domain trained SA2VA-4B~\cite{yuan2025sa2va} and Dense360VLM-3B. Remarkably, ChatGPT4o shows no performance degradation (even slightly higher at 49.33) in the challenging back direction compared to other directions. When trained on the Dense360 dataset, our Dense360VLM-3B outperforms SA2VA-4B in both captioning (51.78 vs 47.80) and grounding (76.81 vs 74.39). Furthermore, Dense360VLM mitigates the performance gap in back-direction captioning, benefiting from our specially designed ERP-RoPE for ERP image inputs.

\subsection{Ablation Study and Analysis}
\label{sec:ablation}
To validate the effectiveness of our Dense360 Dataset and ERP-RoPE, we conducted comprehensive ablation studies as shown in Tab.~\ref{tab:main_exp}. When using the full dataset, integrating ERP-RoPE significantly enhances the MLLM's capability to understand panoramic scenes. Specifically, through ERP-RoPE integration, Dense360VLM achieves a 5.92-point improvement in captioning performance and a 16.38-point boost in grounding accuracy. Under identical ERP-RoPE integration conditions, utilizing the complete Dense360 data yields the highest scores for captioning and grounding tasks, demonstrating that our data components do not exhibit mutual suppression.

\begin{table}[t!]
    \centering
    \caption{Benchmark results. MLLM\dag ~denotes the MLLM that has undergone a second round of post-training using our Dense360 dataset.}
    \label{tab:main_exp}
    \renewcommand{\arraystretch}{1.2}
    \resizebox{0.85\textwidth}{!}{
    \begin{tabular}{c|ccccc|ccccc}
    \toprule[1.5pt]
         \multirow{2}{*}{Method} & \multicolumn{5}{c|}{Captioning} & \multicolumn{5}{c}{Grounding} \\
         ~ & Front & Right & Back & Left & Omnidirection &  Front & Right & Back & Left & Omnidirection \\
         \midrule[1pt]
         Qwen2.5VL-72B~\cite{bai2025qwen2_5vl} & 32.15 & 33.10 & 33.05 & 32.06 & 32.44 & 36.70 & 39.43  & 26.91 & 38.30 & 35.89 \\
         InternVL3-78B~\cite{zhu2025internvl3} & 28.95 & 29.06 & 32.00 & 29.63 & 29.58 & 55.36 & 51.74 & 46.06 & 49.13 & 50.50 \\
         ChatGPT4o-latest & 45.06 & 47.03 & 48.33 & 46.04 & 46.42 & - & - & - & - & - \\
         SA2VA-4B~\cite{yuan2025sa2va}~\dag & 49.89 & 49.49 & 42.98 & 49.68 & 47.80 & 74.66 & 74.12 & 74.25 & 73.93 & 74.39 \\
        \hline 
         Dense360VLM-3B & 52.67 & 53.21 & 50.45 & 52.47 & 51.78 & 76.83 & 76.73 & 76.56 & 76.70 & 76.81 \\
    \bottomrule[1.5pt]
    \end{tabular}
    }
\end{table}

\begin{table}[t!]
    \centering
    \vspace{-5mm}
    \caption{The effectiveness of our Dense360 dataset and ERP-RoPE. Dense360-Caption denotes dense entity-level captions, Dense360-RefSeg refers to unique referring expressions, and Dense360-GCG represents entity-grounded panoramic scene descriptions.}
    \label{tab:main_ablation}
    \renewcommand{\arraystretch}{1.2}
    \resizebox{0.85\textwidth}{!}{
    \begin{tabular}{ccccc|cc|cc}
    \toprule[1.5pt]
        \multicolumn{5}{c|}{Data \& Architecture} & \multicolumn{2}{c|}{Captioning} & \multicolumn{2}{c}{Grounding}\\
        Baseline & Dense360-Caption & Dense360-RefSeg & Dense360-GCG & ERP-RoPE & Back & Omnidirection & Back & Omnidirection \\
        \midrule[1pt]
        \textcolor{merit_red}{\cmark} & \textcolor{merit_gray}{\xmark} & \textcolor{merit_gray}{\xmark} & \textcolor{merit_gray}{\xmark} & \textcolor{merit_gray}{\xmark} & 14.67 & 14.63 & - & - \\
        \textcolor{merit_red}{\cmark} & \textcolor{merit_red}{\cmark} & \textcolor{merit_gray}{\xmark} & \textcolor{merit_gray}{\xmark} & \textcolor{merit_red}{\cmark} & 48.86 & 49.75 & - & - \\
        \textcolor{merit_red}{\cmark} & \textcolor{merit_gray}{\xmark} & \textcolor{merit_red}{\cmark} & \textcolor{merit_gray}{\xmark} & \textcolor{merit_red}{\cmark} & - & - & 70.28 & 70.28 \\
        \textcolor{merit_red}{\cmark} & \textcolor{merit_gray}{\xmark} & \textcolor{merit_gray}{\xmark} & \textcolor{merit_red}{\cmark} & \textcolor{merit_red}{\cmark} & - & - & 46.32 & 43.65 \\
        \textcolor{merit_red}{\cmark} & \textcolor{merit_red}{\cmark} & \textcolor{merit_red}{\cmark} & \textcolor{merit_gray}{\xmark} & \textcolor{merit_red}{\cmark} & 47.78 & 49.41 & 70.03 & 69.88 \\
        \textcolor{merit_red}{\cmark} & \textcolor{merit_red}{\cmark} & \textcolor{merit_gray}{\xmark} & \textcolor{merit_red}{\cmark} & \textcolor{merit_red}{\cmark} & 50.16 & 50.85 & - & - \\
        \textcolor{merit_red}{\cmark} & \textcolor{merit_gray}{\xmark} & \textcolor{merit_red}{\cmark} & \textcolor{merit_red}{\cmark} & \textcolor{merit_red}{\cmark} & - & - & 70.01 & 70.52 \\
        \textcolor{merit_red}{\cmark} & \textcolor{merit_red}{\cmark} & \textcolor{merit_red}{\cmark} & \textcolor{merit_red}{\cmark} & \textcolor{merit_gray}{\xmark} & 40.54 & 45.86 & 69.42 & 60.43 \\
        \textcolor{merit_red}{\cmark} & \textcolor{merit_red}{\cmark}  & \textcolor{merit_red}{\cmark} & \textcolor{merit_red}{\cmark} & \textcolor{merit_red}{\cmark} & 50.45 & 51.78 & 76.56 & 76.81 \\
    \bottomrule[1.5pt]
    \end{tabular}
    }
\end{table}
\section{Conclusion}
\label{sec:conclusion}

In this work, we propose an omnidirectional panoramas dataset for dense visual-language understanding in panoramic settings. We automatically generate a series of reliability-scored annotations through a three-tier pipeline, including 5M dense entity-level captions, 1M unique referring expressions, and 100K entity-grounded panoramic scene descriptions. Our proposed ERP-RoPE effectively addresses two key challenges in ERP images:
i) spatial continuity along the circle of latitude, and
ii) latitude-dependent variation in information density.
This work aims to inspire the community to advance toward comprehensive omnidirectional panoramic scene understanding.

\textbf{Limitations and Future Work.} 
%In this work, we only evaluate the fundamental capabilities of MLLMs in panoramic settings for captioning and grounding, while a more comprehensive evaluation system remains to be developed. We train an expert model for omnidirectional panoramas; in the future, we plan to empower Dense360VLM with support for additional types of visual input.
\dz{While this study focuses on evaluating the core capabilities of MLLMs in panoramic captioning and grounding, a more comprehensive benchmarking framework remains an important direction for future work. In addition, a'l the proposed Dense360VLM is currently designed for ERP panoramas, we plan to extend its compatibility to broader forms of visual input.}

\clearpage
\appendix

% \section{Appendix For More Experiments}
% \label{app_sec:more_exp}

\section{Appendix For More Dataset Examples}
\label{app_sec:more_examples}

\begin{figure}[h]
  \centering
  \includegraphics[width=1.0\textwidth]{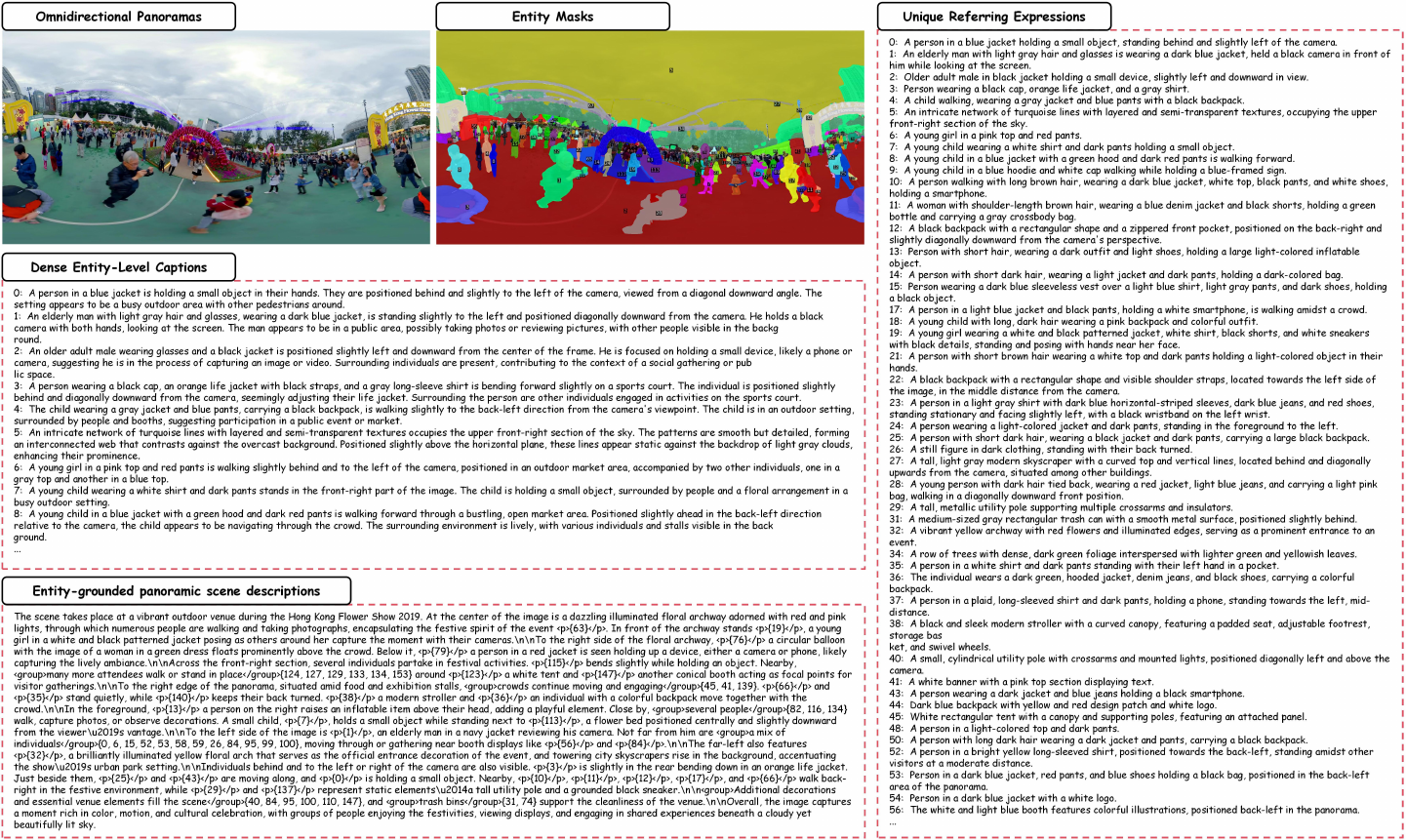}
  \includegraphics[width=1.0\textwidth]{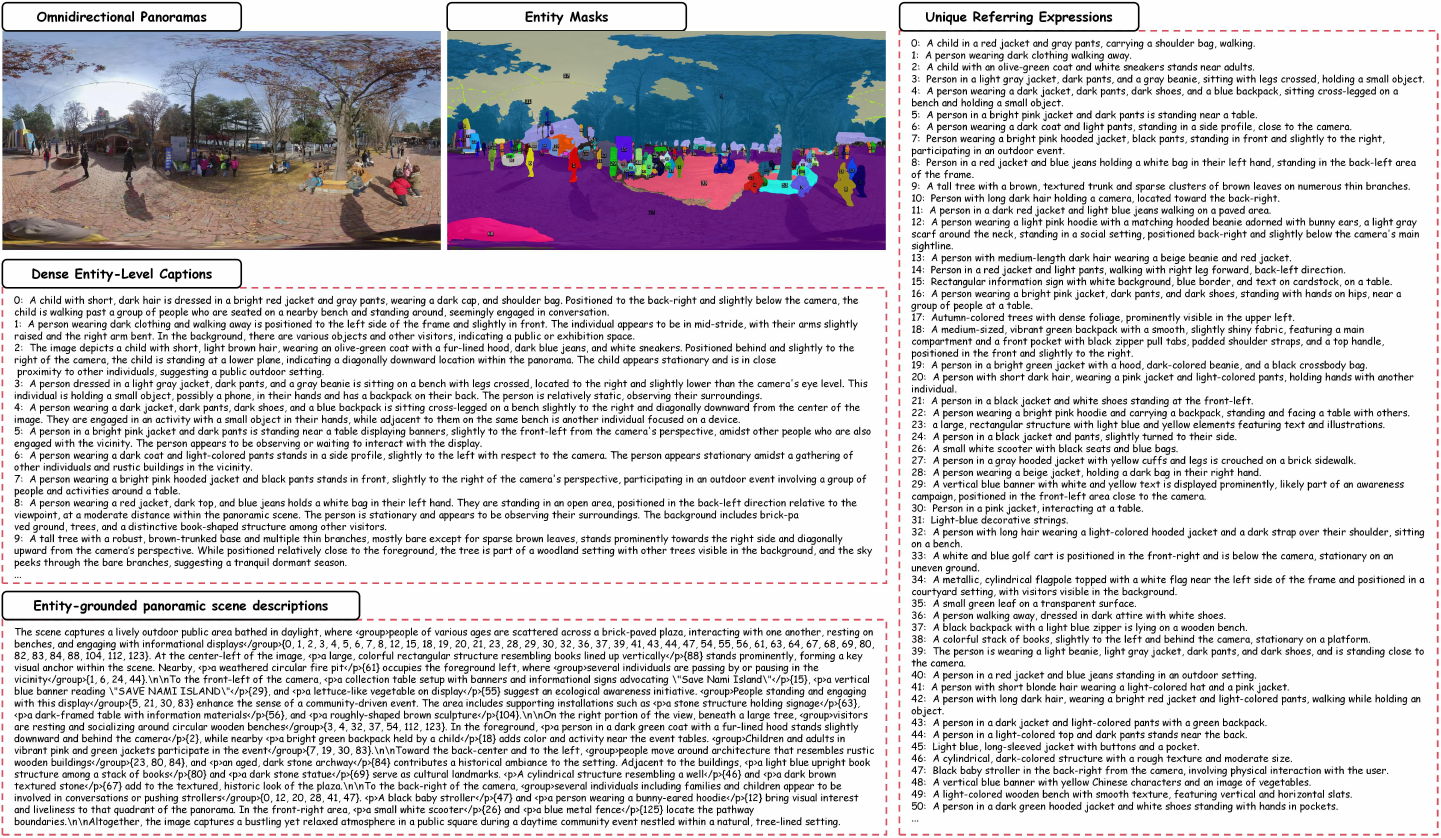}
  \caption{More Examples from the Dense360 Dataset. We employ equirectangular projection (ERP) to represent omnidirectional panoramas. For each ERP image, there exists an entity-grounded panoramic scene description. The entities within ERP images are annotated with dense entity-level captions and unique referring expressions. }
  \label{fig:data_example}
\end{figure}

\begin{figure}[t]
  \centering
  \includegraphics[width=1.0\textwidth]{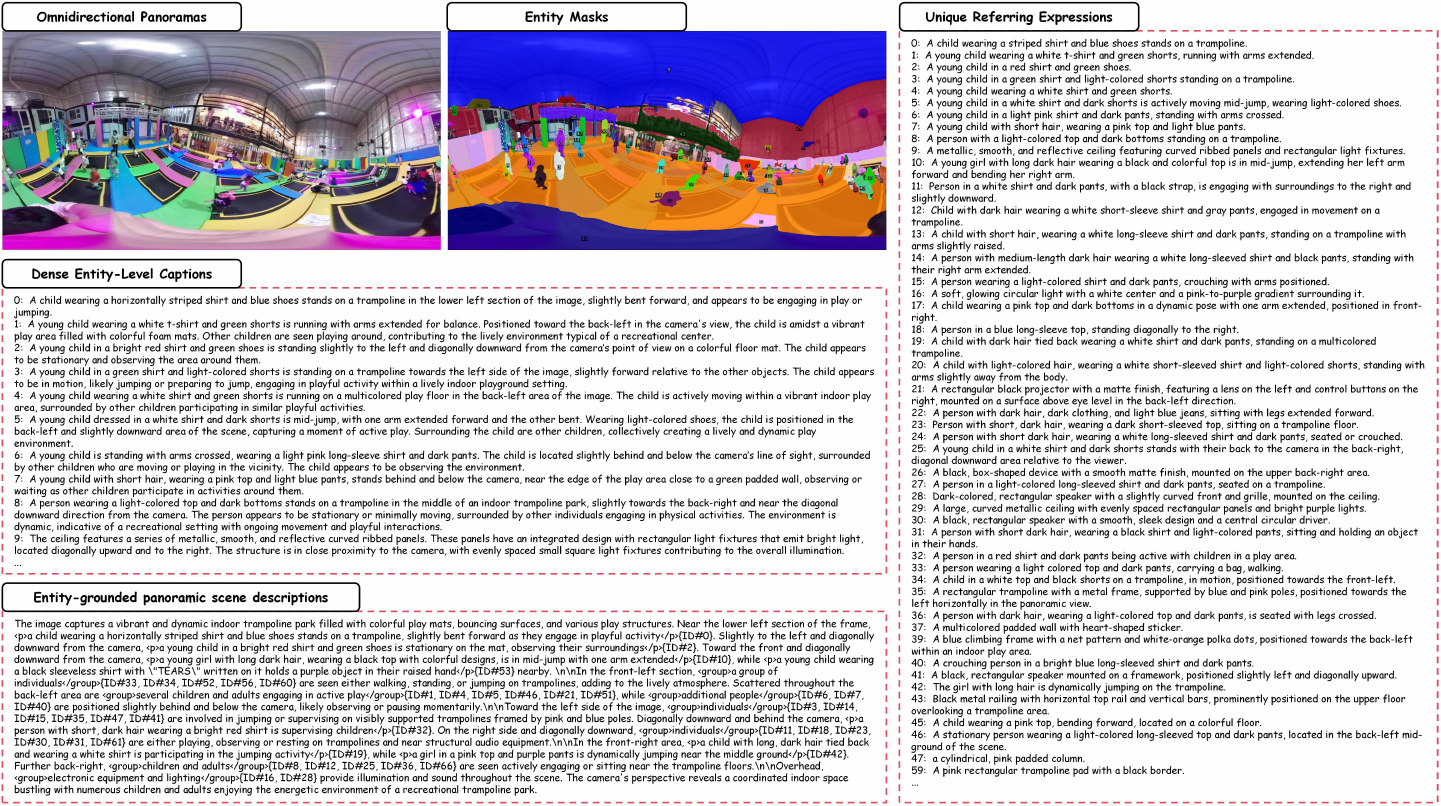}
  \includegraphics[width=1.0\textwidth]{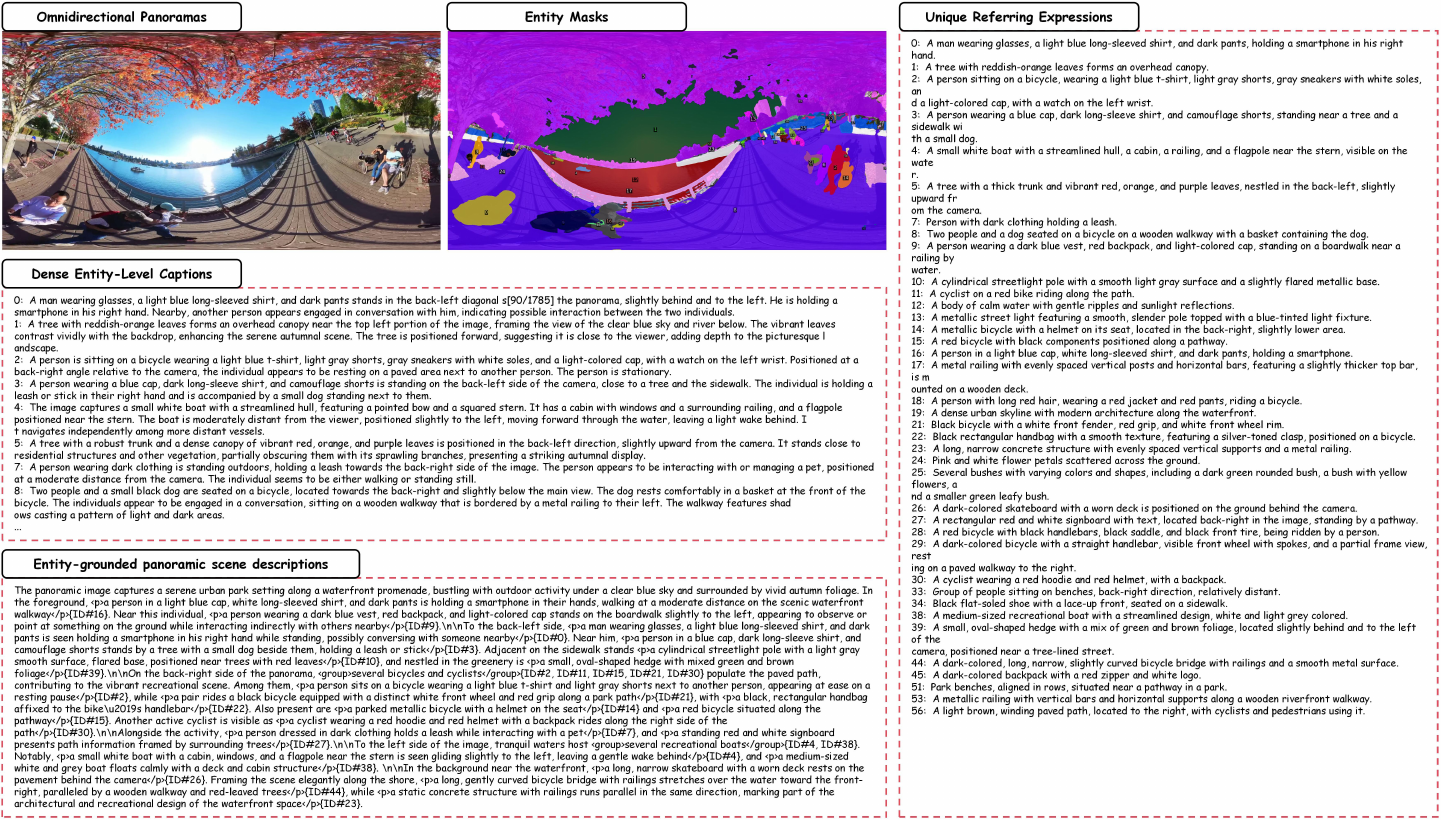}
  \caption{More Examples from the Dense360 Dataset. We employ equirectangular projection (ERP) to represent omnidirectional panoramas. For each ERP image, there exists an entity-grounded panoramic scene description. The entities within ERP images are annotated with dense entity-level captions and unique referring expressions. }
  \label{fig:data_example}
\end{figure}

\clearpage
\newpage
{
    \small
    \bibliographystyle{ieeenat_fullname}
    \bibliography{neurips_2025}

\begin{thebibliography}{10}\itemsep=-1pt

\bibitem{alayrac2022flamingo}
Jean-Baptiste Alayrac, Jeff Donahue, Pauline Luc, Antoine Miech, Iain Barr, Yana Hasson, Karel Lenc, Arthur Mensch, Katherine Millican, Malcolm Reynolds, et~al.
\newblock Flamingo: a visual language model for few-shot learning.
\newblock In {\em NeurIPS}, 2022.

\bibitem{bai2025qwen2_5vl}
Shuai Bai, Keqin Chen, Xuejing Liu, Jialin Wang, Wenbin Ge, Sibo Song, Kai Dang, Peng Wang, Shijie Wang, Jun Tang, et~al.
\newblock Qwen2. 5-vl technical report.
\newblock {\em arXiv preprint arXiv:2502.13923}, 2025.

\bibitem{bhandari2020egok360}
Keshav Bhandari, Mario~A DeLaGarza, Ziliang Zong, Hugo Latapie, and Yan Yan.
\newblock Egok360: A 360 egocentric kinetic human activity video dataset.
\newblock In {\em ICIP}, 2020.

\bibitem{brown2020language}
Tom Brown, Benjamin Mann, Nick Ryder, Melanie Subbiah, Jared~D Kaplan, Prafulla Dhariwal, Arvind Neelakantan, Pranav Shyam, Girish Sastry, Amanda Askell, et~al.
\newblock Language models are few-shot learners.
\newblock In {\em NeurIPS}, 2020.

\bibitem{cai2024vip}
Mu Cai, Haotian Liu, Siva~Karthik Mustikovela, Gregory~P Meyer, Yuning Chai, Dennis Park, and Yong~Jae Lee.
\newblock Vip-llava: Making large multimodal models understand arbitrary visual prompts.
\newblock In {\em CVPR}, 2024.

\bibitem{chatterjee2023opening}
Dibyadip Chatterjee, Fadime Sener, Shugao Ma, and Angela Yao.
\newblock Opening the vocabulary of egocentric actions.
\newblock In {\em NeurIPS}, 2023.

\bibitem{chen2024360+}
Hao Chen, Yuqi Hou, Chenyuan Qu, Irene Testini, Xiaohan Hong, and Jianbo Jiao.
\newblock 360+ x: A panoptic multi-modal scene understanding dataset.
\newblock In {\em CVPR}, 2024.

\bibitem{chen2024sharegpt4v}
Lin Chen, Jinsong Li, Xiaoyi Dong, Pan Zhang, Conghui He, Jiaqi Wang, Feng Zhao, and Dahua Lin.
\newblock Sharegpt4v: Improving large multi-modal models with better captions.
\newblock In {\em ECCV}, 2024.

\bibitem{chen2024single}
Yangyi Chen, Xingyao Wang, Hao Peng, and Heng Ji.
\newblock A single transformer for scalable vision-language modeling.
\newblock In {\em TMLR}, 2024.

\bibitem{chen2024expanding}
Zhe Chen, Weiyun Wang, Yue Cao, Yangzhou Liu, Zhangwei Gao, Erfei Cui, Jinguo Zhu, Shenglong Ye, Hao Tian, Zhaoyang Liu, et~al.
\newblock Expanding performance boundaries of open-source multimodal models with model, data, and test-time scaling.
\newblock {\em arXiv preprint arXiv:2412.05271}, 2024.

\bibitem{chen2024internvl}
Zhe Chen, Jiannan Wu, Wenhai Wang, Weijie Su, Guo Chen, Sen Xing, Muyan Zhong, Qinglong Zhang, Xizhou Zhu, Lewei Lu, et~al.
\newblock Internvl: Scaling up vision foundation models and aligning for generic visual-linguistic tasks.
\newblock In {\em CVPR}, 2024.

\bibitem{chrisley2003embodied}
Ron Chrisley.
\newblock Embodied artificial intelligence.
\newblock In {\em Artificial intelligence}, 2003.

\bibitem{2023xtuner}
XTuner Contributors.
\newblock Xtuner: A toolkit for efficiently fine-tuning llm.
\newblock \url{https://github.com/InternLM/xtuner}, 2023.

\bibitem{dai2023instructblip}
Wenliang Dai, Junnan Li, Dongxu Li, Anthony Meng~Huat Tiong, Junqi Zhao, Weisheng Wang, Boyang Li, Pascale Fung, and Steven Hoi.
\newblock Instructblip: Towards general-purpose vision-language models with instruction tuning.
\newblock {\em arXiv preprint arXiv:2305.06500}, 2023.

\bibitem{devlin2019bert}
Jacob Devlin, Ming-Wei Chang, Kenton Lee, and Kristina Toutanova.
\newblock Bert: Pre-training of deep bidirectional transformers for language understanding.
\newblock In {\em NAACL}, 2019.

\bibitem{diao2024unveiling}
Haiwen Diao, Yufeng Cui, Xiaotong Li, Yueze Wang, Huchuan Lu, and Xinlong Wang.
\newblock Unveiling encoder-free vision-language models.
\newblock {\em arXiv preprint arXiv:2406.11832}, 2024.

\bibitem{diao2025evev2}
Haiwen Diao, Xiaotong Li, Yufeng Cui, Yueze Wang, Haoge Deng, Ting Pan, Wenxuan Wang, Huchuan Lu, and Xinlong Wang.
\newblock Evev2: Improved baselines for encoder-free vision-language models.
\newblock {\em arXiv preprint arXiv:2502.06788}, 2025.

\bibitem{ding2025pvuw}
Henghui Ding, Chang Liu, Nikhila Ravi, Shuting He, Yunchao Wei, Song Bai, Philip Torr, Kehuan Song, Xinglin Xie, Kexin Zhang, et~al.
\newblock Pvuw 2025 challenge report: Advances in pixel-level understanding of complex videos in the wild.
\newblock {\em arXiv preprint arXiv:2504.11326}, 2025.

\bibitem{dosovitskiy2020image}
Alexey Dosovitskiy, Lucas Beyer, Alexander Kolesnikov, Dirk Weissenborn, Xiaohua Zhai, Thomas Unterthiner, Mostafa Dehghani, Matthias Minderer, Georg Heigold, Sylvain Gelly, et~al.
\newblock An image is worth 16x16 words: Transformers for image recognition at scale.
\newblock {\em arXiv preprint arXiv:2010.11929}, 2020.

\bibitem{fan2024embodied}
Yue Fan, Xiaojian Ma, Rongpeng Su, Jun Guo, Rujie Wu, Xi Chen, and Qing Li.
\newblock Embodied videoagent: Persistent memory from egocentric videos and embodied sensors enables dynamic scene understanding.
\newblock {\em arXiv preprint arXiv:2501.00358}, 2024.

\bibitem{fei2025path}
Hao Fei, Yuan Zhou, Juncheng Li, Xiangtai Li, Qingshan Xu, Bobo Li, Shengqiong Wu, Yaoting Wang, Junbao Zhou, Jiahao Meng, et~al.
\newblock On path to multimodal generalist: General-level and general-bench.
\newblock {\em arXiv preprint arXiv:2505.04620}, 2025.

\bibitem{fu2025scene}
Rao Fu, Jingyu Liu, Xilun Chen, Yixin Nie, and Wenhan Xiong.
\newblock Scene-llm: Extending language model for 3d visual reasoning.
\newblock In {\em WACV}, 2025.

\bibitem{han2024free}
Kai Han, Jianyuan Guo, Yehui Tang, Wei He, Enhua Wu, and Yunhe Wang.
\newblock Free video-llm: Prompt-guided visual perception for efficient training-free video llms.
\newblock {\em arXiv preprint arXiv:2410.10441}, 2024.

\bibitem{hu2021lora}
Edward~J Hu, Yelong Shen, Phillip Wallis, Zeyuan Allen-Zhu, Yuanzhi Li, Shean Wang, Lu Wang, and Weizhu Chen.
\newblock Lora: Low-rank adaptation of large language models.
\newblock {\em arXiv preprint arXiv:2106.09685}, 2021.

\bibitem{huang2023open}
Xinyu Huang, Yi-Jie Huang, Youcai Zhang, Weiwei Tian, Rui Feng, Yuejie Zhang, Yanchun Xie, Yaqian Li, and Lei Zhang.
\newblock Open-set image tagging with multi-grained text supervision.
\newblock {\em arXiv preprint arXiv:2310.15200}, 2023.

\bibitem{huang2025egocentric}
Yifei Huang, Jilan Xu, Baoqi Pei, Yuping He, Guo Chen, Mingfang Zhang, Lijin Yang, Zheng Nie, Jinyao Liu, Guoshun Fan, et~al.
\newblock An egocentric vision-language model based portable real-time smart assistant.
\newblock {\em arXiv preprint arXiv:2503.04250}, 2025.

\bibitem{jia2021scaling}
Chao Jia, Yinfei Yang, Ye Xia, Yi-Ting Chen, Zarana Parekh, Hieu Pham, Quoc Le, Yun-Hsuan Sung, Zhen Li, and Tom Duerig.
\newblock Scaling up visual and vision-language representation learning with noisy text supervision.
\newblock In {\em ICML}, 2021.

\bibitem{kundu2025probres}
Sanjoy Kundu, Shanmukha Vellamchetti, and Sathyanarayanan~N Aakur.
\newblock Probres: Probabilistic jump diffusion for open-world egocentric activity recognition.
\newblock {\em arXiv preprint arXiv:2504.03948}, 2025.

\bibitem{lai2024lisa}
Xin Lai, Zhuotao Tian, Yukang Chen, Yanwei Li, Yuhui Yuan, Shu Liu, and Jiaya Jia.
\newblock Lisa: Reasoning segmentation via large language model.
\newblock In {\em CVPR}, 2024.

\bibitem{le2024jrdb}
Duy~Tho Le, Chenhui Gou, Stavya Datta, Hengcan Shi, Ian Reid, Jianfei Cai, and Hamid Rezatofighi.
\newblock Jrdb-panotrack: An open-world panoptic segmentation and tracking robotic dataset in crowded human environments.
\newblock In {\em CVPR}, 2024.

\bibitem{lee2025perspective}
Phillip~Y Lee, Jihyeon Je, Chanho Park, Mikaela~Angelina Uy, Leonidas Guibas, and Minhyuk Sung.
\newblock Perspective-aware reasoning in vision-language models via mental imagery simulation.
\newblock {\em arXiv preprint arXiv:2504.17207}, 2025.

\bibitem{lescop2017360}
Laurent Lescop.
\newblock 360 vision, from panoramas to vr.
\newblock In {\em Envisioning architecture: space/time/meaning}, 2017.

\bibitem{li2024llava}
Bo Li, Yuanhan Zhang, Dong Guo, Renrui Zhang, Feng Li, Hao Zhang, Kaichen Zhang, Peiyuan Zhang, Yanwei Li, Ziwei Liu, et~al.
\newblock Llava-onevision: Easy visual task transfer.
\newblock {\em arXiv preprint arXiv:2408.03326}, 2024.

\bibitem{li2023blip}
Junnan Li, Dongxu Li, Silvio Savarese, and Steven Hoi.
\newblock Blip-2: Bootstrapping language-image pre-training with frozen image encoders and large language models.
\newblock In {\em ICML}, 2023.

\bibitem{li2022blip}
Junnan Li, Dongxu Li, Caiming Xiong, and Steven Hoi.
\newblock Blip: Bootstrapping language-image pre-training for unified vision-language understanding and generation.
\newblock In {\em ICML}, 2022.

\bibitem{li2024bevformer}
Zhiqi Li, Wenhai Wang, Hongyang Li, Enze Xie, Chonghao Sima, Tong Lu, Qiao Yu, and Jifeng Dai.
\newblock Bevformer: learning bird's-eye-view representation from lidar-camera via spatiotemporal transformers.
\newblock In {\em TPAMI}, 2024.

\bibitem{lian2025describe}
Long Lian, Yifan Ding, Yunhao Ge, Sifei Liu, Hanzi Mao, Boyi Li, Marco Pavone, Ming-Yu Liu, Trevor Darrell, Adam Yala, et~al.
\newblock Describe anything: Detailed localized image and video captioning.
\newblock {\em arXiv preprint arXiv:2504.16072}, 2025.

\bibitem{liao2022kitti}
Yiyi Liao, Jun Xie, and Andreas Geiger.
\newblock Kitti-360: A novel dataset and benchmarks for urban scene understanding in 2d and 3d.
\newblock In {\em TPAMI}, 2022.

\bibitem{lim2025ureca}
Sangbeom Lim, Junwan Kim, Heeji Yoon, Jaewoo Jung, and Seungryong Kim.
\newblock Ureca: Unique region caption anything.
\newblock {\em arXiv preprint arXiv:2504.05305}, 2025.

\bibitem{lin2022egocentric}
Kevin~Qinghong Lin, Jinpeng Wang, Mattia Soldan, Michael Wray, Rui Yan, Eric~Z Xu, Difei Gao, Rong-Cheng Tu, Wenzhe Zhao, Weijie Kong, et~al.
\newblock Egocentric video-language pretraining.
\newblock In {\em NeurIPS}, 2022.

\bibitem{liu2024llava1.5}
Haotian Liu, Chunyuan Li, Yuheng Li, and Yong~Jae Lee.
\newblock Improved baselines with visual instruction tuning.
\newblock {\em arXiv preprint arXiv:2310.03744}, 2024.

\bibitem{liu2024improved}
Haotian Liu, Chunyuan Li, Yuheng Li, and Yong~Jae Lee.
\newblock Improved baselines with visual instruction tuning.
\newblock In {\em CVPR}, pages 26296--26306, 2024.

\bibitem{Liu2023llava}
Haotian Liu, Chunyuan Li, Qingyang Wu, and Yong~Jae Lee.
\newblock Visual instruction tuning.
\newblock {\em arXiv preprint arXiv:2304.08485}, 2023.

\bibitem{liu2023visual}
Haotian Liu, Chunyuan Li, Qingyang Wu, and Yong~Jae Lee.
\newblock Visual instruction tuning.
\newblock In {\em NeurIPS}, 2023.

\bibitem{liu2025seg}
Yuqi Liu, Bohao Peng, Zhisheng Zhong, Zihao Yue, Fanbin Lu, Bei Yu, and Jiaya Jia.
\newblock Seg-zero: Reasoning-chain guided segmentation via cognitive reinforcement.
\newblock {\em arXiv preprint arXiv:2503.06520}, 2025.

\bibitem{liu2024vmamba}
Yue Liu, Yunjie Tian, Yuzhong Zhao, Hongtian Yu, Lingxi Xie, Yaowei Wang, Qixiang Ye, Jianbin Jiao, and Yunfan Liu.
\newblock Vmamba: Visual state space model.
\newblock In {\em NeurIPS}, 2024.

\bibitem{luo2024mono}
Gen Luo, Xue Yang, Wenhan Dou, Zhaokai Wang, Jiawen Liu, Jifeng Dai, Yu Qiao, and Xizhou Zhu.
\newblock Mono-internvl: Pushing the boundaries of monolithic multimodal large language models with endogenous visual pre-training.
\newblock In {\em CVPR}, 2025.

\bibitem{luo2024feast}
Gen Luo, Yiyi Zhou, Yuxin Zhang, Xiawu Zheng, Xiaoshuai Sun, and Rongrong Ji.
\newblock Feast your eyes: Mixture-of-resolution adaptation for multimodal large language models.
\newblock {\em arXiv preprint arXiv:2403.03003}, 2024.

\bibitem{nie2025wmnav}
Dujun Nie, Xianda Guo, Yiqun Duan, Ruijun Zhang, and Long Chen.
\newblock Wmnav: Integrating vision-language models into world models for object goal navigation.
\newblock {\em arXiv preprint arXiv:2503.02247}, 2025.

\bibitem{o2015introduction}
Keiron O'shea and Ryan Nash.
\newblock An introduction to convolutional neural networks.
\newblock {\em arXiv preprint arXiv:1511.08458}, 2015.

\bibitem{qi2023high}
Lu Qi, Jason Kuen, Tiancheng Shen, Jiuxiang Gu, Wenbo Li, Weidong Guo, Jiaya Jia, Zhe Lin, and Ming-Hsuan Yang.
\newblock High quality entity segmentation.
\newblock In {\em ICCV}, 2023.

\bibitem{radford2021learning}
Alec Radford, Jong~Wook Kim, Chris Hallacy, Aditya Ramesh, Gabriel Goh, Sandhini Agarwal, Girish Sastry, Amanda Askell, Pamela Mishkin, Jack Clark, et~al.
\newblock Learning transferable visual models from natural language supervision.
\newblock In {\em ICML}, 2021.

\bibitem{rang2025eve}
Miao Rang, Zhenni Bi, Chuanjian Liu, Yehui Tang, Kai Han, and Yunhe Wang.
\newblock Eve: Efficient multimodal vision language models with elastic visual experts.
\newblock {\em arXiv preprint arXiv:2501.04322}, 2025.

\bibitem{ravi2024sam}
Nikhila Ravi, Valentin Gabeur, Yuan-Ting Hu, Ronghang Hu, Chaitanya Ryali, Tengyu Ma, Haitham Khedr, Roman R{\"a}dle, Chloe Rolland, Laura Gustafson, et~al.
\newblock Sam 2: Segment anything in images and videos.
\newblock {\em arXiv preprint arXiv:2408.00714}, 2024.

\bibitem{shen2025vlm}
Haozhan Shen, Peng Liu, Jingcheng Li, Chunxin Fang, Yibo Ma, Jiajia Liao, Qiaoli Shen, Zilun Zhang, Kangjia Zhao, Qianqian Zhang, et~al.
\newblock Vlm-r1: A stable and generalizable r1-style large vision-language model.
\newblock {\em arXiv preprint arXiv:2504.07615}, 2025.

\bibitem{shen2024aligning}
Yunhang Shen, Chaoyou Fu, Peixian Chen, Mengdan Zhang, Ke Li, Xing Sun, Yunsheng Wu, Shaohui Lin, and Rongrong Ji.
\newblock Aligning and prompting everything all at once for universal visual perception.
\newblock In {\em CVPR}, 2024.

\bibitem{long-vita}
Yunhang Shen, Chaoyou Fu, Shaoqi Dong, Xiong Wang, Peixian Chen, Mengdan Zhang, Haoyu Cao, Ke Li, Xiawu Zheng, Yan Zhang, et~al.
\newblock Long-vita: Scaling large multi-modal models to 1 million tokens with leading short-context accuray.
\newblock {\em arXiv preprint arXiv:2502.05177}, 2025.

\bibitem{wang2024llm}
Junchi Wang and Lei Ke.
\newblock Llm-seg: Bridging image segmentation and large language model reasoning.
\newblock In {\em CVPR}, 2024.

\bibitem{wang2024qwen2}
Peng Wang, Shuai Bai, Sinan Tan, Shijie Wang, Zhihao Fan, Jinze Bai, Keqin Chen, Xuejing Liu, Jialin Wang, Wenbin Ge, et~al.
\newblock Qwen2-vl: Enhancing vision-language model's perception of the world at any resolution.
\newblock {\em arXiv preprint arXiv:2409.12191}, 2024.

\bibitem{wu2024controlmllm}
Mingrui Wu, Xinyue Cai, Jiayi Ji, Jiale Li, Oucheng Huang, Gen Luo, Hao Fei, Guannan Jiang, Xiaoshuai Sun, and Rongrong Ji.
\newblock Controlmllm: Training-free visual prompt learning for multimodal large language models.
\newblock {\em NeurIPS}, 2024.

\bibitem{yan2024panovos}
Shilin Yan, Xiaohao Xu, Renrui Zhang, Lingyi Hong, Wenchao Chen, Wenqiang Zhang, and Wei Zhang.
\newblock Panovos: Bridging non-panoramic and panoramic views with transformer for video segmentation.
\newblock In {\em ECCV}, 2024.

\bibitem{yang2024qwen2}
An Yang, Baosong Yang, Beichen Zhang, Binyuan Hui, Bo Zheng, Bowen Yu, Chengyuan Li, Dayiheng Liu, Fei Huang, Haoran Wei, et~al.
\newblock Qwen2. 5 technical report.
\newblock {\em arXiv preprint arXiv:2412.15115}, 2024.

\bibitem{yang2023dawn}
Zhengyuan Yang, Linjie Li, Kevin Lin, Jianfeng Wang, Chung-Ching Lin, Zicheng Liu, and Lijuan Wang.
\newblock The dawn of lmms: Preliminary explorations with gpt-4v (ision).
\newblock {\em arXiv preprint arXiv:2309.17421}, 2023.

\bibitem{yang2022lavt}
Zhao Yang, Jiaqi Wang, Yansong Tang, Kai Chen, Hengshuang Zhao, and Philip~HS Torr.
\newblock Lavt: Language-aware vision transformer for referring image segmentation.
\newblock In {\em CVPR}, 2022.

\bibitem{yuan2025sa2va}
Haobo Yuan, Xiangtai Li, Tao Zhang, Zilong Huang, Shilin Xu, Shunping Ji, Yunhai Tong, Lu Qi, Jiashi Feng, and Ming-Hsuan Yang.
\newblock Sa2va: Marrying sam2 with llava for dense grounded understanding of images and videos.
\newblock {\em arXiv preprint arXiv:2501.04001}, 2025.

\bibitem{yuan20254th}
Haobo Yuan, Tao Zhang, Xiangtai Li, Lu Qi, Zilong Huang, Shilin Xu, Jiashi Feng, and Ming-Hsuan Yang.
\newblock 4th pvuw mevis 3rd place report: Sa2va.
\newblock {\em arXiv preprint arXiv:2504.00476}, 2025.

\bibitem{yurtsever2020survey}
Ekim Yurtsever, Jacob Lambert, Alexander Carballo, and Kazuya Takeda.
\newblock A survey of autonomous driving: Common practices and emerging technologies.
\newblock In {\em IEEE access}, 2020.

\bibitem{zeng2023clip2}
Yihan Zeng, Chenhan Jiang, Jiageng Mao, Jianhua Han, Chaoqiang Ye, Qingqiu Huang, Dit-Yan Yeung, Zhen Yang, Xiaodan Liang, and Hang Xu.
\newblock Clip2: Contrastive language-image-point pretraining from real-world point cloud data.
\newblock In {\em CVPR}, 2023.

\bibitem{zhang2025mem2ego}
Lingfeng Zhang, Yuecheng Liu, Zhanguang Zhang, Matin Aghaei, Yaochen Hu, Hongjian Gu, Mohammad~Ali Alomrani, David Gamaliel~Arcos Bravo, Raika Karimi, Atia Hamidizadeh, et~al.
\newblock Mem2ego: Empowering vision-language models with global-to-ego memory for long-horizon embodied navigation.
\newblock {\em arXiv preprint arXiv:2502.14254}, 2025.

\bibitem{zhang2024omg}
Tao Zhang, Xiangtai Li, Hao Fei, Haobo Yuan, Shengqiong Wu, Shunping Ji, Chen~Change Loy, and Shuicheng Yan.
\newblock Omg-llava: Bridging image-level, object-level, pixel-level reasoning and understanding.
\newblock In {\em NeurIPS}, 2024.

\bibitem{zhang2025pixel}
Tao Zhang, Xiangtai Li, Zilong Huang, Yanwei Li, Weixian Lei, Xueqing Deng, Shihao Chen, Shunping Ji, and Jiashi Feng.
\newblock Pixel-sail: Single transformer for pixel-grounded understanding.
\newblock {\em arXiv preprint arXiv:2504.10465}, 2025.

\bibitem{zhang2023dvis}
Tao Zhang, Xingye Tian, Yu Wu, Shunping Ji, Xuebo Wang, Yuan Zhang, and Pengfei Wan.
\newblock Dvis: Decoupled video instance segmentation framework.
\newblock In {\em ICCV}, 2023.

\bibitem{zhang2025dvis++}
Tao Zhang, Xingye Tian, Yikang Zhou, Shunping Ji, Xuebo Wang, Xin Tao, Yuan Zhang, Pengfei Wan, Zhongyuan Wang, and Yu Wu.
\newblock Dvis++: Improved decoupled framework for universal video segmentation.
\newblock {\em IEEE TPAMI}, 2025.

\bibitem{zhang2024seeing}
Xiaofeng Zhang, Yihao Quan, Chaochen Gu, Chen Shen, Xiaosong Yuan, Shaotian Yan, Hao Cheng, Kaijie Wu, and Jieping Ye.
\newblock Seeing clearly by layer two: Enhancing attention heads to alleviate hallucination in lvlms.
\newblock {\em arXiv preprint arXiv:2411.09968}, 2024.

\bibitem{zhang2025enhancing}
Xiaofeng Zhang, Fanshuo Zeng, Yihao Quan, Zheng Hui, and Jiawei Yao.
\newblock Enhancing multimodal large language models complex reason via similarity computation.
\newblock In {\em AAAI}, 2025.

\bibitem{zhong2022regionclip}
Yiwu Zhong, Jianwei Yang, Pengchuan Zhang, Chunyuan Li, Noel Codella, Liunian~Harold Li, Luowei Zhou, Xiyang Dai, Lu Yuan, Yin Li, et~al.
\newblock Regionclip: Region-based language-image pretraining.
\newblock In {\em CVPR}, 2022.

\bibitem{zhou2024improving}
Yikang Zhou, Tao Zhang, Shunping Ji, Shuicheng Yan, and Xiangtai Li.
\newblock Improving video segmentation via dynamic anchor queries.
\newblock In {\em ECCV}, 2024.

\bibitem{zhou2025they}
Yikang Zhou, Tao Zhang, Shilin Xu, Shihao Chen, Qianyu Zhou, Yunhai Tong, Shunping Ji, Jiangning Zhang, Xiangtai Li, and Lu Qi.
\newblock Are they the same? exploring visual correspondence shortcomings of multimodal llms.
\newblock {\em arXiv preprint arXiv:2501.04670}, 2025.

\bibitem{zhu2025internvl3}
Jinguo Zhu, Weiyun Wang, Zhe Chen, Zhaoyang Liu, Shenglong Ye, Lixin Gu, Yuchen Duan, Hao Tian, Weijie Su, Jie Shao, et~al.
\newblock Internvl3: Exploring advanced training and test-time recipes for open-source multimodal models.
\newblock {\em arXiv preprint arXiv:2504.10479}, 2025.

\end{thebibliography}
}

\end{document}